# A Novel Pseudo Nearest Neighbor Classification Method Using Local Harmonic Mean Distance

Junzhuo Chen [1], Zhixin Lu [2, *] *and Shitong Kang* [1]

[1]School of Artificial Intelligence, Hebei University of Technology, Tianjin, 300130, China
[2]College of Big Data, Weifang Institute of Technology, Weifang, 261101, China
*Corresponding Author: Zhixin Lu. Email: l1179484365@163.com

**Abstract:** In the realm of machine learning, the KNN classification algorithm is widely recognized as a straightforward and efficient method for classification. However, its sensitivity to the K value is too high when handling small samples or outliers, which seriously affects its classification performance. The purpose of this article is to introduce a new KNN-based classifier named the Novel Pseudo Nearest Neighbor Classification Method Using Local Harmonic Mean Distance (LMPHNN).LMPHNN is the first to introduce harmonic mean as a measure of distance to improve classification performance based on LMPNN rules and Harmonic mean distance (HMD). For our proposed LMPHNN classifier, we will begin by seeking the k nearest neighbors for every class and produce k distinct local vectors as prototypes. Next, we create pseudo nearest neighbors (PNNs) based on the local mean for each class by comparing the HMD of the sample with the first k group, which is distinct from the initial k group. The classification is ultimately determined by calculating the Euclidean distance between the query sample and PNNs, which is based on the local mean of these categories. To evaluate the effectiveness of the suggested classifier, extensive experiments are conducted on numerous real UCI datasets and combined datasets by employing this approach in addition to seven KNN-based classifiers. We use metrics such as precision, recall, accuracy, and F1 as evaluation criteria. According to the experimental results, LMPHNN achieves an average precision of 97% on all datasets, surpassing other methods by an impressive margin of 14%. The average recall was 80%, a 12% improvement over other methods. The average accuracy of the proposed classifier exceeds that of other classifiers, demonstrating a 5% enhancement. It also achieves a higher average F1 value,13% higher than other methods. In summary, other classifiers are not as good as the proposed classifier in terms of performance, and LMPHNN demonstrates lower sensitivity when dealing with small sample sizes.

Code: *https://github.com/jzc777/LMPHNN*

**Keywords:** k nearest neighbor; local mean vector; pseudo nearest neighbor; Harmonic mean distance

## 1 Introduction

Classification algorithms are a crucial technique in machine learning. It can automatically determine which data belongs to a category based on the input data, and correctly divide different instances into their respective classifications. Many fields, including text classification and pattern recognition, frequently employ classification algorithms. It possesses the ability to measure and forecast objects effectively, resulting in great usefulness. Currently, the popular classification algorithms in machine learning consist of decision trees, Bayesian classification, support vector machines, neural networks, k nearest neighbor(KNN) algorithms, and so on. The KNN rule [1] is considered a traditional nonparametric classification algorithm [2–4]. Within the realm of pattern classification, Fix and Hodges [5] were the first to introduce the KNN rule, which is known to be one of the simplest nonparametric techniques. The main idea behind the KNN algorithm is to locate the KNNs of the query sample and assign the most representative category to the query sample by majority voting. This algorithm combines classification ideas with simple implementation. KNN is renowned for its excellent performance in the Bayesian sense when the training sample size N and neighborhood size K approach infinity, with the condition that the limit of K/N tends to zero [6]. Despite having numerous advantages, KNN classifiers, there are still some major disadvantages, as follows [7, 8], When dealing with a limited amount of data, particularly when outliers [9] are present, the expected Bayesian asymptotic performance is challenging for the algorithm to attain, KNN ignores the geometric distribution of neighbors and is not always the most suitable when measuring the distance of the query sample from the neighbor, and a suitable distance metric is not found. To mitigate the impact of outliers, Mitani and Hamamoto [10] propose a dependable local mean-based k-nearest neighbor algorithm (LMKNN), in which the local mean vector(LMV) of KNNs of every single class is used to categorize patterns of queries [3, 10]. LMKNN [3] has been in existence since its initial introduction, LMKNN has been effectively utilized in group-based classification [11], discriminant analysis [12], and distance metric learning [13]. Later, To address the issue resulting from selecting KNNs of the identical state, measures need to be taken, Zeng et al. [14] proposed a pseudo-nearest neighbor (PNN) rule using distance-weighted local learning. PNN classifiers are a method of generating pseudo nearest neighbor samples by first finding KNNs samples in each class and then using the distance between different neighbors, instead of using the original nearest neighbor samples. Classification is done by querying the category of the nearest pseudo neighbor. From the extensive experimental results reported in [3, 14], LMKNN and PNN classifiers show significant resistance and superior performance in classification performance. A new PNN classifier is introduced, which utilizes the local mean as a basis(LMPNN) [15]. The LMPNN merges the fundamental concepts of the LMKNN and PNN rules. The LMPNN classifier achieves good classification performance by merging the distance with assigned weights between the query sample and the LMV in each category to obtain the pseudo distance. Although these algorithms are well done for outliers in [3, 10, 14, 16], their sensitivity to small sample sizes remains in terms of classification performance due to sample noise and imprecision [2, 17].

   From the spatial geometric distribution of neighbor locations, the nearest centroid neighbor (NCN) was successfully derived to make neighbors as symmetrical as possible around the query sample position in the event of limited sample size [13]. Subsequently, in the KNN-based pattern classification [18], the first proposal introduces the k-nearest centroid neighbor (KNCN) rule, which builds upon the NCN concept. The main goal of the KNCN rule is to offer a justification for the similarity based on distance and the arrangement in space of k neighbors in the training dataset. Its excellent performance has been demonstrated in numerous experimental studies, especially when dealing with a small sample size.[18–20]. In any case, KNCN overestimates the importance of some centroid neighbors, which can cause classification performance [19–21] to become unreliable, and it is inappropriate for KNCN to assume that k-centroid neighbors have the same weight when making classification decisions. Subsequently, an extended KNCN method was proposed—the LMKNCN classifier [4], the query pattern is assigned to the class label that has the nearest local centroid mean vector, and this scheme comprehensively considers the adjacent relationship and distribution of K neighbors in space, but also takes into account the more dependable local averages within distinct categories when making classification decisions. For the parameter k, it is less sensitive to classification performance. Although the performance of the classifier has surpassed KNCN, it is still at a sufficient level. Nevertheless, it is crucial to acknowledge that it is only appropriate for specific problems. In KNN-based classification, the Euclidean distance is the conventional measure [8, 22, 23]used, which assumes that the data follows a Gaussian isotropic distribution. Conversely, when the size of the neighborhood K is large, the assumption that the community has an isotropic distribution is frequently invalid. Hence, the size of the neighborhood is highly dependent on the community's scale, denoted as K.To determine the sample's distance from the training sample, you need to look for a suitable distance metric, the MLM-KHNN [24] classifier first introduced the HMD as a distance metric, which is different from the traditional Euclidean distance, the HMD metric considers the distance between the local mean under different k values and the query variable x, reducing the dependence and sensitivity to k, LMKHNCN [25] based on this idea also introduced HMD as a distance metric into LMKNCN. Experiments show that the classifier exhibits good results, and since the results obtained using the HMD measure perform well, we use this measure in our proposed method.

Considering the dialogue on appeal, to address the issue of k's sensitivity in the KNN-based classification algorithm and enhance classification performance, we present an LMPHNN classifier. This classifier derives from the LMPNN classifier and employs HMD as a distance metric in addition to the local mean. To decrease sensitivity to k, the initial step of our suggested classifier involves the use of KNNs from every class to create local means. Then, it utilizes the local mean from the top r (r<=k) group in each class to generate local PNNs from the HMD of the query sample. In contrast to the LMPNN classifier, which utilizes Euclidean distance, our proposed approach employs the harmonic mean as a distance metric. While assigning weights to neighbors with different distances from the test point, harmonic means and local means are used to balance the gap between different neighbors, that is, while giving higher weights to closer neighbors, the importance of farther neighbors is considered, and harmonic means are used based on LMPNN to further reduce the sensitivity to K. To assess the accuracy of LMPHNN's classification, we conducted experiments using 9 machine learning datasets obtained from UCI and Kaggle. Experimental findings demonstrate that the suggested approach has a minimal classification error, which affirms its capability as a reliable and efficient classifier.

Statistical pattern recognition researchers have noted that the LMKHNN algorithm's performance is greatly influenced by the selection of a single neighborhood size k for each class, as well as the decision to use a uniform k value across all classes in the rule. However, it is well-known that LMKHNN exhibits robustness to outliers, primarily due to its utilization of the HMD. In contrast, the concept of LMPNN has shown the ability to overcome the influence of outliers, thus leading to improved classification performance.

Taking inspiration from the robustness of LMKHNN and the effectiveness of LMPNN, we propose a novel classification method that aims to combine the strengths of both approaches. This innovative approach is expected to enhance classification performance by mitigating the sensitivity to k values while simultaneously addressing the challenge of outliers.

LMPNN is a PNN classifier based on LMVs, which reduces the effect of noise on the classification results and improves the classification performance by being more accurate in selecting PNNs. Local subclasses are generated by the KNNs of each class. Hence, the selection of parameter k holds significant importance in producing LMVs that accurately portray their respective classes.

Nonetheless, the selection of parameter k in the LMPNN rule presents two main challenges that may lead to misclassifications.

First, selecting an appropriate parameter k plays a crucial role in generating a LMV that effectively encapsulates the characteristics of its corresponding class. Selecting a small value for k might result in the PNN not being accurate enough, thus affecting the classification performance. Selecting a large value for k might result in the PNN containing too much noise, thus affecting the classification performance. Second, if the selected features are not representative enough, then even if the appropriate k value is chosen, it may lead to misclassification. Utilizing the same k value for all classes, as in the LMPNN rule, is not a reasonable approach.

We introduce a novel classifier termed the LMPHNN rule, Using LMPNN's weighted average for all neighbors, leveraging local means to effectively address this issue. The approach involves the computation of the KNNs for each class, resulting in k distinct LMVs. Notably, these LMVs vary in their distances from the query sample, and the ones that have shorter distances are considered more appropriate for representing their respective classes for classification purposes.LVM focuses on local neighborhoods rather than global properties, allowing outliers to be balanced by neighboring normal values. Our emphasis lies in identifying these closer local subclasses, allowing for varying k values across different classes.

To achieve this goal, we propose using HMD, evaluating the similarity between each class's set of multi-LMVs and the query sample x. Subsequently, we classify the sample into the class exhibiting the smallest HMD. In comparison to LMPNN and other KNN-based classifiers, our LMPHNN classifier has a reduced classification error rate and a decreased sensitivity to changes in the neighborhood size k.

The remaining part of our article has been structured in the following manner: In the second section, there is a concise summary of the relevant classifiers and their underlying motivations that have been suggested, which have had an impact on our research. In the third section, we introduce our proposed LMPHNN method and discuss its application in various characteristic factors. In the fourth section, we perform a large number of experimental comparisons between the real UCI dataset and the Kaggle dataset, validate our method with other competing KNN-based classifiers, and demonstrate the superiority of our method over these competing KNN-based classifiers. Section 5 of our discussion delved into the necessary complexity for computing a rival classifier, and we provided an overall summary of the entire article in Section 6.

## 2 Structure

One of the top nonparametric classifiers in text classification algorithms is the KNN algorithm. There have been numerous KNN-based optimization algorithms developed in the past few years. Examples include algorithms such as LMKNN, which is founded upon the KNN algorithm using local mean, and LMPNN, which is built upon pseudo nearest neighbor using the local mean classifier. In this section, we will examine the impact of LMKNN and LMPNN classifiers on our work, as well as the concept of utilizing harmonic means as a distance metric in this article.

### *2.1 The LMKNN rule*

Mitani and Hamamoto [3] introduced LMKNN as a solution to counteract the impact of outliers present in the training set (TS), and experiments have proved that LMKNN can serve outliers well in small-sample training. LMKNN relies on Euclidean distance as a means to determine classification decisions. To be more specific, for every category within the training set, we identify the KNNs to the query point x, and subsequently calculate the LMV of these k neighbors. Finally, the classification judgment is made according to the Euclidean distance between the query point x and these LMVs. In general classification problems, it is supposed a training set $TS = \{x_n \in \mathbb{R}^d\}_{n=1}^{N}$ is the TS eigenspace given d dimension, where N represents the overall count of training samples, and $y_n \in \{c_1, c_2, \dots, c_M\}$ for the class label of $x_n$. $TSc_i = \{x_j^i \in \mathbb{R}^m\}_{j=1}^{N_i}$ represents a subset of the class $c_I$, $N_I$ denotes the overall quantity of training examples for the class. In LMKNN rules, the query pattern x can be determined using these steps.

(1)    Retrieve the $c_I$ nearest neighbors from the set associated with each class k(k<=$N_i$) of the query sample, and look for the query pattern x.

Let $T_{\omega_i}^k(x) = \{x_j^i \in \mathbb{R}^d\}_{j=1}^{K}$ be the set of KNNs of x in class $c_I$ using the Euclidean distance metric, and the Euclidean distance is utilized for calculating the distance from x to y..e. Eq(1)

$$d(x, x_j^i) = \sqrt{(x - x_j^i)^T (x - x_j^i)} \qquad (1)$$

(2)    Determine the LMV $lm_{c_i}^k$ of the local values obtained from the set $T_{ik}^{NN}(x)$ of class $c_i$:

$$\mathrm{lm}_{c_i}^k = \frac{1}{k}\sum\nolimits_{j=1}^{k} x_j^i \tag{2}$$

(3) Apply equation(1)to determine the distance $d(x, \mathrm{lm}_{c_i}^k)$ separating the test point x and the LMV $\mathrm{lm}_{c_i}^k$ of class $c_i$.

(4) If point x represents the test point that has the least distance to the LMV of the class $c_I$ compared to other classes, it is classified as belonging to the class $c_i$.

$$c = \arg\min\nolimits_{c_i} d(x, \mathrm{lm}_{c_i}^k) \tag{3}$$

It is important to note that when k = 1, the LMKNN and 1-NN classifiers yield the same classification result. However, in LMKNN, the concept of K is distinct from that of KNN. KNN selects k nearest neighbors from the entire training dataset, while LMKNN utilizes the local mean vectors of k nearest neighbors within every single category. The objective of LMKNN is to identify the category that is most similar to the local area of the query sample. As a result, incorporating local mean vectors helps to effectively mitigate the impact of outliers, particularly when dealing with limited sample sizes.

## 2.2 The LMPNN rule

To enhance the classification performance of the LMKNN classifier, further improvements are needed, Gou [15] et al. proposed a pseudo-nearest neighbor classifier (LMPNN) based on the local mean. This classifier's design draws inspiration from both the PNN[14] and the LMKNN[3]. Experimental results indicate that the classifier exhibits strong classification performance. To begin with the LMPNN classifier, the initial step involves identifying the set of test points as well as the KNNs from each class and then finding the local mean of the neighbors of r group (r<=k) in each set, and then the sum of the weighted distances is computed by assigning weights to each local mean to form the PNN of the local mean. The classification label is then determined by comparing the distance sizes of various local mean pseudo-nearest neighbors. Based on the observation of the extensive experimental results in the range [4, 22], it can be concluded that the LMPNN classifier is more stable in dealing with outliers and has better classification performance than the traditional KNN classifier.

When provided with a query point x and a training set $TS = \{x_n \in \mathbb{R}^d\}_{n=1}^{N}$, $TSc_i(x) = \{x_j^i \in \mathbb{R}^d\}_{j=1}^{N_i}$ is a subset of TS in class $c_i$, where d is the characteristic dimension, N and $N_i$ are the sample sizes of $TS$ and $TS_{c_i}^k$, separately. To determine the class label of x in the LMPNN rule, the following steps are followed:

(1)Find the KNNs of the test point x from $TS_{c_i}$ of each class $c_I$ in the training set TS, such as $TS_{c_i}^k(x) = \{x_j^i \in \mathbb{R}^d\}_{j=1}^{K}$, And the KNNs $x_1^i, x_2^i \ldots \ldots x_k^i$ are order blue in ascending order according to their Euclidean distance from the query point x.

(2)Calculate the LMV $\bar{x}_j^i$ of the first j(j<=k) nearest neighbors of the test point x in class $c_i$

$$\bar{x}_j^i = \frac{1}{j}\sum\nolimits_{r=1}^{j} x_r^i \tag{4}$$

Let $\bar{TS}_{c_i}^k(x) = \{\bar{x}_j^i \in \mathbb{R}^d\}_{j=1}^{k}$ represent the set of k LMVs corresponding to KNNs in class $c_i$ , $d(x, \bar{x}_1^i), d(x, \bar{x}_2^i), \ldots, d(x, \bar{x}_k^i)$ is the corresponding Euclidean distance from $\bar{x}_j^i$ to x in the set $\bar{TS}_{c_i}^k(x)$

(3)Give varying weights to the k LMVs of each class, the weight $\bar{W}_j^i$ of the j th LMV $\bar{x}_j^i$ obtained from class $c_i$ is determined as:

$$\bar{W}_j^i = \frac{1}{j} \quad j = 1, \dots, k \tag{5}$$

(4) Determine the local mean-based PNN of the test point x for each class, Let $\bar{x}_i^{PNN}$ represent the local mean-based PNN of x in the class $c_i$. The distance $d(x, x_i^{PNN})$ from x to $x_i^{PNN}$ is calculated as follows:

$$d(x, \bar{x}_i^{PNN}) = \left(\bar{W}_1^i \times d(x, \bar{x}_1^i) + \bar{W}_2^i \times d(x, \bar{x}_2^i) + \cdots + \bar{W}_k^i \times d(x, \bar{x}_k^i)\right) \tag{6}$$

(5) According to the local mean of Equation (6), class $c_i$ is assigned to the test point x, which is where the pseudo nearest neighbor is situated:

$$c = \arg\min_{c_i} d(x, \bar{x}_i^{PNN}) \tag{7}$$

*2.3 Harmonic Mean Distance*

The KNN classifier uses Euclidean distance as the metric and assumes data follows a Gaussian isotropic distribution. Nonetheless, when the number of chosen neighbors k is large, the assumption of a Gaussian isotropic distribution is frequently unsuitable. Consequently, the sensitivity of KNN to the number of neighbors increases significantly. The MLM-KHNN [24] classifier first utilizes HMD as a similarity measure. The main emphasis is on calculating the reliability of local means within each category, thus prioritizing more dependable methods across different categories. Unlike the LMKNN classifier, this method reduces the error rate by reducing the sensitivity to the parameter k, rather than using Euclidean distances to unify the k values of all k means in each class like the LMKNN classifier. The reason for incorporating the HMD metric into our proposed method is due to its strong performance. To calculate the HMD metric, one must sum the harmonic mean of the Euclidean distances between a particular data point and another set of data points. In this paper, we use HMD to generate local mean pseudo nearest neighbors of LMPNN. When the traditional LMPNN generates the PNNs of the local mean, the sum of the weights of the top j(j<=k) local mean and the Euclidean distance of the query sample is replaced by the HMD, and the sum of the local mean weights of the top j(j<=k) group and the harmonic mean of the query sample in every single class are solved, and the local pseudo nearest neighbor is obtained, and In the end, the test sample is categorized as the point with the lowest local average. For instance, if x represents a query sample, it is supposed $\bar{TS}_{c_i}^k(x) = \{\bar{x}_j^i \in \mathbb{R}^d\}_{r=1}^{j}$ represents a vector in the set that corresponds to the first nearest j(j<=k) LMV in the class $c_I$, $\bar{x}_j^i = \frac{1}{j}\sum_{r=1}^{j} x_r^I$ represents the LMV of the test point x and the nearest neighbors of the first j(j<=k) in the ith class.

Afterwards, HMD $\left(x, \{\bar{x}_j^i\}_{r=1}^{j}\right)$ from the test point x and set $\bar{TS}_{c_i}^k(x)$ is calculated as:

$$\text{HMD}\left(x, \{\bar{x}_j^i\}_{r=1}^{j}\right) = \frac{j}{\sum_{r=1}^{j} \frac{1}{d\left(x, \{\bar{x}_j^i\}_{r=1}^{j}\right)}} \tag{8}$$

The use of HMD makes the increase or decrease of a small number of k values (usually the neighbors near the edge of the class) in the reciprocal case have a minimal impact on the performance of the algorithm. By synthesizing the distance between multiple local mean vectors rather than a single k value, the sensitivity to k values is effectively reduced. The specific impact on k values will be discussed in detail later in this paper. (HMD is more responsive to real relationships between classes and can reduce dependence on any single distance measure, so the detection of small samples is also significantly improved in performance)

Because larger values of edge anomalies are given less weight when taking the reciprocal, robustness to outliers (which are usually far from the query point) is improved. At the same time, for reasonable samples, HMD provides a weight determined by distance to balance the importance of neighbors in the class, which gives a better balance in the case of uneven distribution or containing outliers.

In this process, other distance measurement methods have serious defects: Euclidean distance emphasizes "distance" too much in higher dimensional feature space, and is too sensitive to outliers; The calculation complexity of Chebyshev distance is high. Manhattan distance is limited by the grid layout and so on.

## 3 The presented LMPHNN classifier

### 3.1 Description of PNN rules

A brief overview of the PNN's core code will facilitate the reader's understanding of the following sections:

1. Calculate distance: For each sample in the test set, calculate the distance between it and all samples in the training set.

2. Find the nearest neighbors: For each test sample, according to the calculated distance, find its k nearest neighbors.

3. Generate pseudo nearest neighbors: Use different strategies (such as distance weighted or unweighted) to generate pseudo nearest neighbors from k nearest neighbors. If distance weighting is used, the pseudo-nearest neighbor is more affected by neighbors with smaller distances.

4. Classification decision: According to the category distribution of pseudo-nearest neighbors, a category label is assigned to the test sample through majority voting or calculation of weighted distance.

5. Output predictions: Finally, a predicted category label is provided for each sample in the test set.

In the LMPHNN model, after identifying the k nearest neighbors of a single category of the query sample, we capture the local features of this category by calculating the local top r (r<=k) group mean vector of the k nearest neighbors, and then by querying the harmonic average distance of the sample (considering the relationship between the local mean and the query sample, The pseudo-nearest neighbor is created based on the local mean of each class, and finally the class of the query sample is determined by calculating the Euclidean distance between the sample and the pseudo-nearest neighbor.

It can be seen from the above description that selecting top r (r<=k) different local vectors for PNN to capture and represent the local features and class structure of their respective classes can improve the robustness of the model even when the sample size is small and there are outliers, which is of high significance.

### *3.2 The LMPHNN classification rule*

The LMPHNN is presented in this subsection as an extension of the LMPNN rule. Although the method we propose has a formal resemblance to the LMPNN rule, it is fundamentally different from it in terms of the scheme for finding PNNs. In the LMPNN rule, the LMVs of the first KNNs in each category are first computed. Then, the PNNs are determined by using the k LMVs of these categories. The LMPHNN method, on the other hand, introduces the HMD between multiple LMVs in each category and the query sample $x$ is measured to determine the similarity between them. Ultimately, the samples are categorized based on the categories with pseudo-nearest HMDs.

Let $T = \{x_n \in \mathbb{R}^d\}_{n=1}^N$ consist of a set of $M$ classes for training purposes $\omega_1, \ldots, \omega_M$ and $T_{\omega_i} = \{x_j^i \in \mathbb{R}^d\}_{j=1}^{N_i}$ be a class subset of $T$ from the class $\omega_i$, where d represents the feature dimension, $N$ and $N_i$ represent the sample numbers of $T$ and $T_{\omega_i}$, respectively. In the proposed LMPHNN rule, the class label of a query point x is determined through the following steps:

(1) Find the KNNs from $T_{\omega_i}$ for each class in $\omega_i$ or the test point x in the training set $T$, referred to as $T_{\omega_i}^k(x) = \{x_j^i \in \mathbb{R}^d\}_{j=1}^k$, and the KNNs $x_1^i, x_2^i, \ldots, x_k^i$ are sorted in ascending order based on their corresponding Euclidean distances to $x$.

(2) Calculate the LMV $\bar{x}_j^i$ of the first $j$ nearest neighbors of a test point $x$ from class $\omega_i$.

$$\bar{x}_j^i(x) = \frac{1}{j}\sum_{l=1}^j x_l^i \qquad (9)$$

Let $\bar{T}_{\omega_i}^k(x) = \{\bar{x}_j^i \in \mathbb{R}^d\}_{j=1}^k$ denote the set of the $k$ LMVs corresponding to the KNNs in the class $\omega_i$, and $d(x, \bar{x}_1^i), d(x, \bar{x}_2^i), \ldots, d(x, \bar{x}_k^i)$ represent the Euclidean distances to $x$ for each of them. $HMD(x, \bar{x}_j^k)$ is the HMD between $x$ and the $k$ multi-LMVs $\bar{x}_j^i$

$$HMD(x, \bar{x}_j^k) = \frac{k}{\sum_{i=1}^k \frac{1}{d(x,\bar{x}_j^i)}} \qquad (10)$$

Note that the LMV $\bar{x}_1^i$ for the first nearest neighbor, denoted as $x_1^j$, is identical to this neighbor.

(3) Allocate varying weights to the $k$ LMVs in each class, following a similar approach as PNN. Specifically, the weight $\bar{W}_j^i$ of the $j-th$ LMV $\bar{x}_j^i$ from the class $\omega_i$ is calculated as:

$$\bar{W}_j^i = \frac{1}{j} \quad j = 1, \ldots, k \qquad (11)$$

(4) Discover the LMPNN for the sample point $x$ within every class. Denote $\bar{\chi}_i^{PNN}$ as the LMPNN for $x$ from class $\omega_i$, and let $\omega_i$ represent the class label of $\omega_i$ The harmonic distance $HMD(x, \bar{x}_i^{PNN})$ between $x$ and $\bar{\chi}_i^{PNN}$ is computed as follows:

$$HMD(x, \bar{x}_i^{PNN}) = \bar{W}_1^i \times HMD(x, \bar{x}_1^i) + \bar{W}_2^i \times HMD(x, \bar{x}_2^i) + \cdots + \bar{W}_k^i \times HMD(x, \bar{x}_k^i) \qquad (12)$$

(5) Classify the test point $x$ as belonging to class c based on the LMPNN distance metric, which is the HMD to the nearest neighbors, as defined by Eq. (12), of all the classes.

$$c = \arg\min_{\omega_i} HMD(x, \bar{x}_i^{PNN}) \qquad (13)$$

### 3.3 The LMPHNN algorithm
The steps to implement the LMPHNN classifier are as follows:
**Requirements:**

$x$: a query pattern

$T = \{x_n \in \mathbb{R}^d\}_{n=1}^{N}$: a training set

$N_1, \ldots, N_M$: the amount of training samples for each of the $M$ classes.

$\omega_1, \ldots, \omega_M$: $M$ class labels.

$T_{\omega_i} = \{x_j^i \in \mathbb{R}^d\}_{j=1}^{N_i}$: a training subset from the class $\omega_i$

$k$: the neighborhood size.

$M$: the number of classes in $T$

**Ensure**: Predict the label assigned to a query pattern for classification purposes based on its closest LMPNN among classes.

***Step 1***: Compute the distances from class $\omega_i$ to $x$ for each of the training samples.

**for** $j = 1$ to $N_i$ **do**

$$d(x, x_j^i) = \sqrt{(x - x_j^i)^T (x - x_j^i)}$$

**end for**

***Step 2***: Find the KNN of $x$ in $T_{\omega i}$, sorted in ascending order based on their distances $d(x, x_j^i)$, say $T_{\omega_i}^k(x) = \{x_j^i \in \mathbb{R}^d\}_{j=1}^{k}$

***Step 3***: Using $T_{\omega_i}^k(x)$, Compute the LMV $\bar{x}_j^i$ of $x$'s first $j$ nearest neighbors, and then calculate the HMD $\text{HMD}(x, \bar{x}_j^k)$ between $x$ and $k$ multi-LMVs $\bar{x}_j^i$.

**for** $j = 1$ to $k$ **do**

$$\bar{x}_j^i(x) = \frac{1}{j} \sum_{l=1}^{j} x_l^i$$

$$d(x, \bar{x}_j^i) = \sqrt{(x - \bar{x}_j^i)^T (x - \bar{x}_j^i)}$$

$$\text{HMD}(x, \bar{x}_j^k) = \frac{k}{\sum_{i=1}^{k} \frac{1}{d(x, \bar{x}_j^i)}}$$

**end for**

Subsequently, establish $\bar{T}_{\omega_i}^k(x) = \{\bar{x}_j^i \in \mathbb{R}^d\}_{j=1}^{k}$ and $\bar{D}_{\omega_i}^k = \{\text{HMD}(x, \bar{x}_1^i), \ldots, \text{HMD}(x, \bar{x}_k^i)\}$

***Step 4***: Allocate the weight $\bar{W}_j^i$ to the $j$-th LMV $\bar{x}_j^i$ in the set $T_{\omega_i}^k(X)$.

**for** $j = 1$ to $k$ **do**

$\bar{W}_j^i = \frac{1}{j}$  $j = 1, \ldots, k$

**end for**

Subsequently, set $\bar{W}_{\omega_i}^k = \{\bar{W}_1^i, \ldots, \bar{W}_k^i\}$

***Step 5***: Using $\bar{W}_{\omega_i}^k$ and $\bar{D}_{\omega_i}^k$, finding the LMPNN $\bar{\chi}_i^{PNN}$ that satisfies the given constraint

$$HMD(x, \bar{x}_i^{\text{PNN}}) = \bar{W}_1^i \times HMD(x, \bar{x}_1^i) + \bar{W}_2^i \times HMD(x, \bar{x}_2^i) + \cdots + \bar{W}_k^i \times HMD(x, \bar{x}_k^i)$$

**Step 6**: Assign the class $c$ of the LMPNN with the nearest HMD to $x$.

$$c = \arg\min_{\omega_i} HMD(x, \bar{x}_i^{PNN})$$

Note that when $k = 1$, it is categorized into LMPNN rules and PNN rules, which exhibit comparable classification performance to the 1-NN classifier. Section 3.2 provides a detailed explanation of the pseudo-code used for the LMPHNN classifier.

### 3.4 Difference between LMPHNN and LMPNN

In pattern classification, computational complexity plays a crucial role in designing efficient classifiers. To elucidate the advantages of the proposed LMPHNN classifier, in this section, we will examine and compare the computational complexity of the LMPNN and LMPHNN classifiers. Specifically, our focus is on assessing the computational complexity associated with online computations during the classification phase.

In the following discussion, the following notation is utilized by us: $n$ for the overall quantity of training examples, $n_I$ for the number of training examples belonging to the class $\omega_i$, $d$ for the feature space is characterized by the number of dimensions it has, and $c$ for the number of classes.

For the LMPNN classifier:

In the stage of classification, the initial step involves searching for the KNN in each class by using Euclidean distance as the determining factor. The computational complexity of this operation can be described as $O(n_1 d + n_2 d + \cdots + n_c d)$, which can also be abbreviated as $O(nd)$. Moreover, the $O(n_1 k + n_2 k + \cdots + n_c k)$ comparisons remain unchanged, which are equal to $O(nk)$.

Subsequently, we compute $k$ LMVs corresponding to the KNN per class and determine the distances between these categorical $k$ LMVs and the query pattern.

The step's computational complexity can be described as $O(2ckd)$ is required. Next, we assign weight $w_j$ to the LMVs per class and find the $j-$ th LMPNN for each class. This step requires computational operations on the order of $O(3ck)$.

The final step involves determining the class of the query pattern using a certain approach, denoted as $O(c)$. Consequently, the computational complexity of LMPNN is $O(nd + nk + 2ckd + 3ck + c)$ in all its steps.

For the proposed LMPHNN classifier:

There are four steps involved in the classification process of the proposed LMPHNN classifier. The initial step is similar to the LMPNN rule, where LMPNN initially computes the distances between the query pattern and all the training samples of each class with a complexity of $O(nd + nk)$.

The LMPHNN method acquires $k$ multilocalized mean vectors during the second step, which requires the use of $O(2ckd)$.

For each class, the D is computed between the query $x$ and $k$ multi-LMVs in the third step. This calculation involves $O(cdk)$ multiplications and $O(cdk + ck)$ sum operations, as demonstrated in Eq(9).

Fourth step, the weight $W_j$ is designated to the $j-$ th LMV per class, and the LMPNN for each class is determined. This step also demands $O(3ck)$.

In conclusion, the proposed method determines the class with the lowest HMD to the given query and assigns the query sample to that class through comparisons, which $O(c)$ represents computational complexity. Therefore, the overall computational complexity of the LMPHNN rule can be summarized as $O(nd + nk + 4cdk + 4ck + c)$.

## 4 Experiments

To confirm the classification performance of the proposed LMPHNN, we conduct a comparison between LMPHNN and other competing classifiers: KNN[1], LMKNN[3], KNCN[18], LMPNN[15], LMKHNN[24], LMKNCN[4] and PNN[14]. Extensive experiments on 9 real UCI and Kaggle datasets were conducted to study the error rate, which is a highly effective metric in the domain of pattern classification. In addition, we verify the effectiveness of the proposed approach on various real datasets by assessing additional metrics including accuracy, recall, precision, and F1.

Both UCI and Kaggle datasets come from real scenarios with a huge amount of data. In terms of data set features, differences in sample size, attribute number, and category number can extensively test the performance of the classifier in handling different tasks. Strong complexity also requires high robustness of the classifier itself. When parameter k is selected, the selection range is wide, and the LMPHNN classifier can be evaluated fairly and comprehensively.

All the experiments of the study were conducted on @2.30GHz 11th generation Intel Core i7-11800H CPU. The Windows 10 operating system is running on a system memory of 16gb, with a 64-bit architecture. The experimental platform was Matlab R2020a.

### 4.1 Datasets

In this subsection, we offer a summary of the datasets that we used in our experiments. A total of nine datasets from UCI and Kaggle machine learning datasets are selected, which are sourced from real-world scenarios. These 9 real-world datasets differ significantly in terms of sample size, attributes, and categories.

Its main features include sample size, attributes, classes and test sets are displayed in the table provided below.

**Table 1:** Datasets from UCI and Kaggle Used in Experiments

|  | Samples | Attributes | Classes | Testing set |
|---|---|---|---|---|
| Predictive Maintenance(PM) | 10000 | 12 | 2 | 5000 |
| Wine | 178 | 13 | 3 | 79 |
| Sleep health and lifestyle(SL) | 378 | 12 | 3 | 189 |
| Titanic | 889 | 8 | 3 | 444 |
| Airline Passenger Satisfaction(APS) | 981 | 23 | 3 | 490 |
| Bank Marketing(BM) | 4878 | 17 | 2 | 2439 |
| Breast Cancer(BC) | 4024 | 16 | 2 | 2012 |
| Telecom Churn(TC) | 2666 | 20 | 2 | 1333 |
| Milk Quality Prediction(MQP) | 1059 | 8 | 3 | 529 |

### 4.2 Experiments on real UCI, Kaggle datasets

To demonstrate and objectively emphasise the classification performance of the proposed LMPHNN

method, the LMPHNN is experimentally compared with KNN, LMKNN, LMKNCN, KNCN, LMPNN, and PNN on nine real UCI, Kaggle datasets through error rates.

For classifying query samples, the LMKNN rule uses LMVs in each class. To achieve improved classification performance, PNN and LMPNN rules are successfully developed and built upon the LMKNN rule. The PNN rule initially seeks to acquire PNNs within each class and subsequently assigns the labels of the nearest PNNs to the query samples, while the LMPNN rule is a synthesis of the concepts from both the LMKNN and the PNN approaches. KNCN is a classification algorithm whose rule uses each class of the KNN samples to classify the query samples. LMKNCN is an improved version of the KNCN algorithm that computes the LMV of the nearest neighbor samples. LMKHNN is a measure of local similarity based on LMKNN that uses HMD.

**Table 2:** Accuracy of each model

| Accuracy | LMPHNN | PNN | LMPNN | LMKNN | LMKHNN | KNN | KNCN | LMKNCN |
|---|---|---|---|---|---|---|---|---|
| PM | **0.5294** | 0.5208 | 0.5266 | 0.5241 | 0.5278 | 0.5066 | 0.5184 | 0.5209 |
| Wine | **0.7890** | 0.7191 | 0.7341 | 0.7553 | 0.7665 | 0.6704 | 0.7079 | 0.7054 |
| SL | **0.8507** | 0.7808 | 0.8333 | 0.8363 | 0.8130 | 0.8315 | 0.7945 | 0.8459 |
| Titanic | **0.7585** | 0.6236 | 0.7370 | 0.7055 | 0.7447 | 0.6929 | 0.6987 | 0.7067 |
| APS | **0.6009** | 0.4299 | 0.5748 | 0.5755 | 0.5957 | 0.5741 | 0.5571 | 0.5506 |
| BM | **0.7732** | 0.6281 | 0.7580 | 0.7463 | 0.7676 | 0.7413 | 0.7217 | 0.7252 |
| BC | **0.8857** | 0.8258 | 0.8703 | 0.8576 | 0.8781 | 0.8710 | 0.8442 | 0.8462 |
| TC | **0.8649** | 0.1620 | 0.8456 | 0.8250 | 0.8571 | 0.8080 | 0.8097 | 0.8110 |
| MQP | **0.9966** | 0.9401 | 0.9945 | 0.9836 | 0.9943 | 0.9861 | 0.9910 | 0.9962 |
|  | **0.7832** | 0.6256 | 0.7638 | 0.7566 | 0.7716 | 0.7424 | 0.7381 | 0.7453 | 0.7408 |

**Table 3:** Recall of each model

| Recall | LMPHNN | PNN | LMPNN | LMKNN | LMKHNN | KNN | KNCN | LMKNCN |
|---|---|---|---|---|---|---|---|---|
| PM | 0.5655 | 0.5572 | 0.5629 | 0.5593 | 0.5636 | 0.3632 | 0.5959 | **0.6062** |
| Wine | **0.8084** | 0.7440 | 0.7579 | 0.7759 | 0.7909 | 0.6179 | 0.5659 | 0.4976 |
| SL | **0.8658** | 0.8003 | 0.8512 | 0.8502 | 0.8315 | 0.8083 | 0.7675 | 0.8151 |
| Titanic | **0.7815** | 0.6531 | 0.7625 | 0.7319 | 0.7707 | 0.6369 | 0.2219 | 0.2315 |
| APS | **0.6344** | 0.4640 | 0.6105 | 0.6082 | 0.6294 | 0.3915 | 0.0102 | 0.0163 |
| BM | **0.7972** | 0.6639 | 0.7827 | 0.7708 | 0.7913 | 0.6638 | 0.6765 | 0.7009 |
| BC | 0.8981 | 0.8433 | 0.8846 | 0.8741 | 0.8915 | 0.6875 | 0.9067 | **0.9193** |
| TC | **0.8820** | 0.2532 | 0.8624 | 0.8466 | 0.8751 | 0.6981 | 0.2864 | 0.2509 |
| MQP | **0.9971** | 0.9468 | 0.9952 | 0.9845 | 0.9950 | 0.9802 | 0.9781 | 0.9907 |

**Table 4:** Precision of each model

| Precision | LMPHNN | PNN | LMPNN | LMKNN | LMKHNN | KNN | KNCN | LMKNCN |
|---|---|---|---|---|---|---|---|---|
| PM | 0.9310 | 0.9299 | 0.9303 | **0.9316** | 0.9310 | 0.5111 | 0.5992 | 0.5996 |
| Wine | **0.9742** | 0.9618 | 0.9659 | 0.9703 | 0.9665 | 0.6464 | 0.8632 | 0.9658 |
| SL | 0.9816 | 0.9712 | 0.9771 | **0.9827** | 0.9766 | 0.7921 | 0.8866 | 0.9290 |
| Titanic | **0.9687** | 0.9448 | 0.9636 | 0.9586 | 0.9642 | 0.6290 | 0.3584 | 0.3606 |
| APS | **0.9401** | 0.8990 | 0.9331 | 0.9372 | 0.9386 | 0.4219 | 0.0748 | 0.0920 |
| BM | 0.9681 | 0.9531 | 0.9670 | 0.9662 | **0.9687** | 0.7813 | 0.6989 | 0.6948 |
| BC | **0.9861** | 0.9788 | 0.9833 | 0.9803 | 0.9846 | 0.7444 | 0.9091 | 0.9011 |
| TC | 0.9801 | 0.9049 | **0.9801** | 0.9726 | 0.9793 | 0.5793 | 0.3253 | 0.3139 |
| MQP | 0.9996 | 0.9928 | 0.9994 | 0.9992 | 0.9994 | 0.9847 | **1.0000** | **1.0000** |

**Table 5:** F1 of each model

| F1 | LMPHNN | PNN | LMPNN | LMKNN | LMKHNN | KNN | KNCN | LMKNCN |
|---|---|---|---|---|---|---|---|---|
| PM | **0.6923** | 0.6849 | 0.6899 | 0.6877 | 0.6909 | 0.4196 | 0.5975 | 0.6029 |
| Wine | **0.8820** | 0.8366 | 0.8466 | 0.8603 | 0.8678 | 0.6297 | 0.6833 | 0.6564 |
| SL | **0.9193** | 0.8765 | 0.9090 | 0.9108 | 0.8969 | 0.7996 | 0.8226 | 0.8682 |
| Titanic | **0.8627** | 0.7681 | 0.8486 | 0.8272 | 0.8537 | 0.6242 | 0.2738 | 0.2816 |
| APS | **0.7507** | 0.6013 | 0.7300 | 0.7305 | 0.7466 | 0.4049 | 0.0179 | 0.0276 |
| BM | **0.8720** | 0.7716 | 0.8623 | 0.8546 | 0.8685 | 0.7092 | 0.6875 | 0.6978 |
| BC | **0.9394** | 0.9046 | 0.9306 | 0.9233 | 0.9351 | 0.7040 | 0.9079 | 0.9101 |
| TC | **0.9275** | 0.2789 | 0.9163 | 0.9039 | 0.9230 | 0.6297 | 0.3046 | 0.2786 |
| MQP | **0.9983** | 0.9691 | 0.9973 | 0.9917 | 0.9972 | 0.9820 | 0.9889 | 0.9953 |

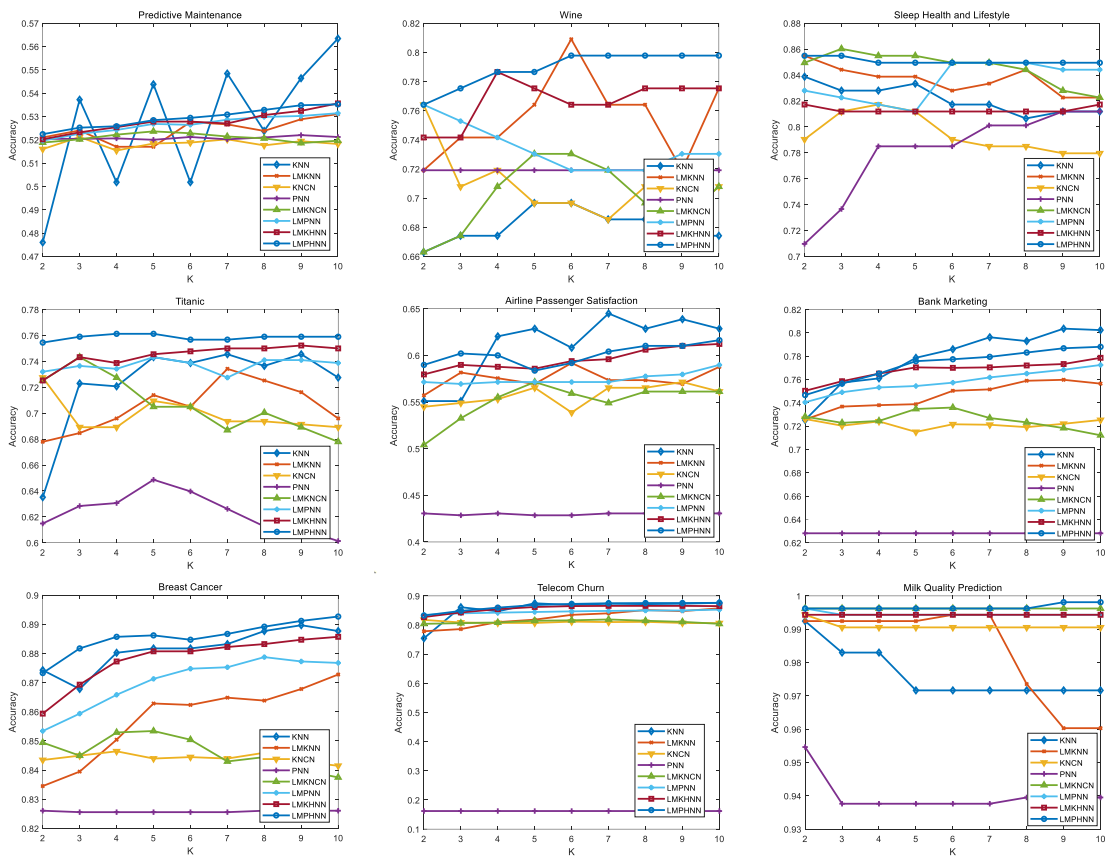

**Figure 1:** Accuracy curve of each model on the datasets

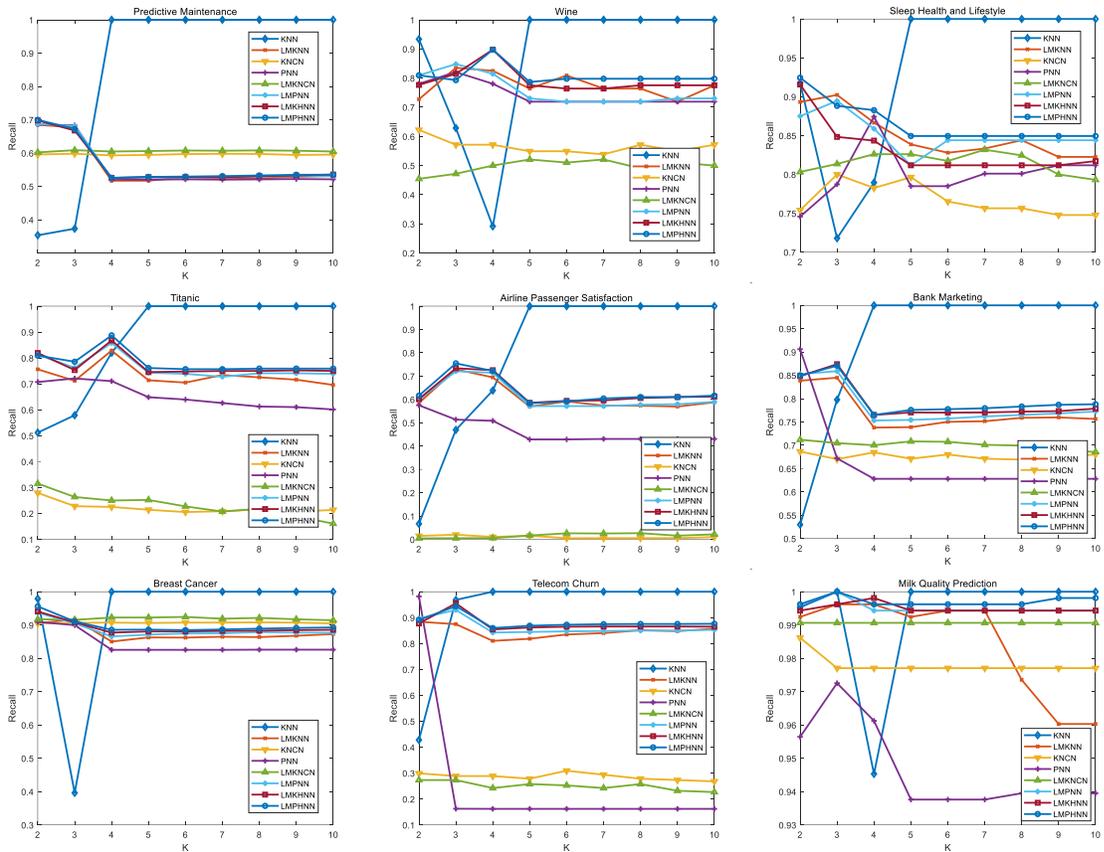

**Figure 2:** Recall accuracy curve of each model on the datasets

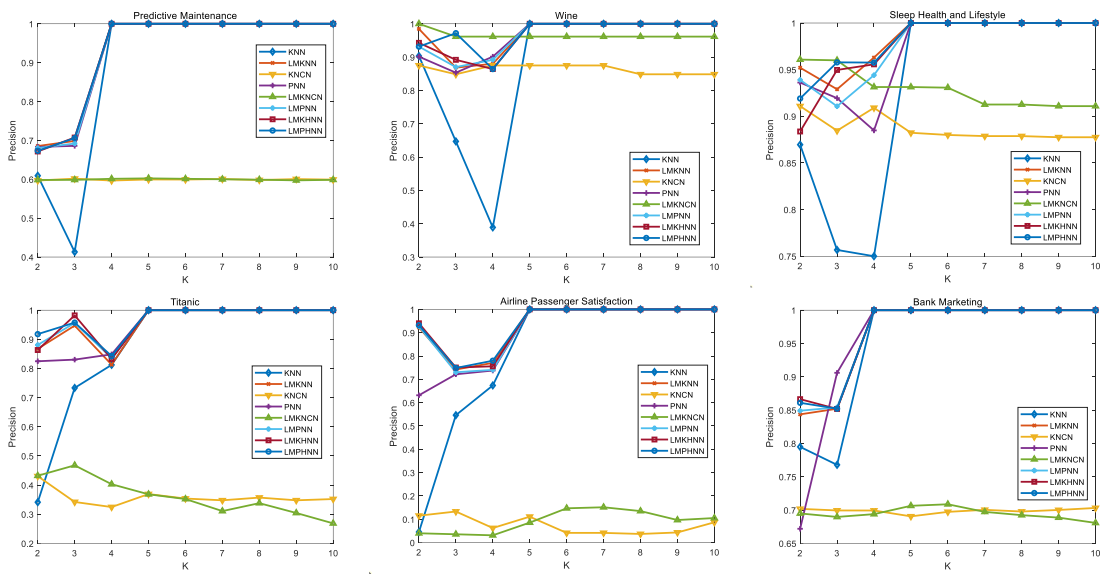

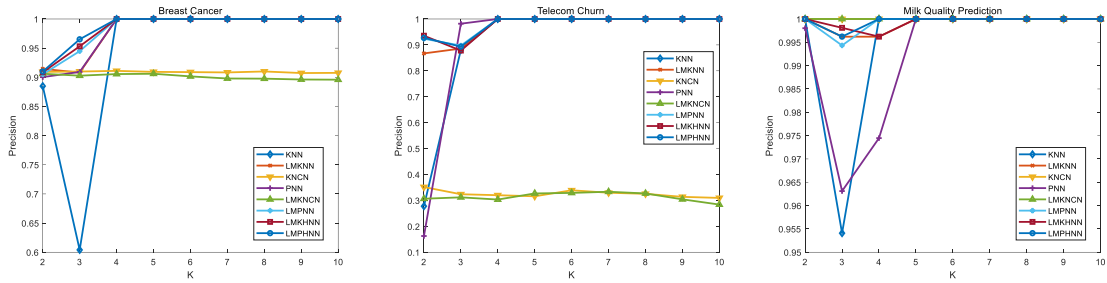

**Figure 3:** Precision accuracy curve of each model on the datasets

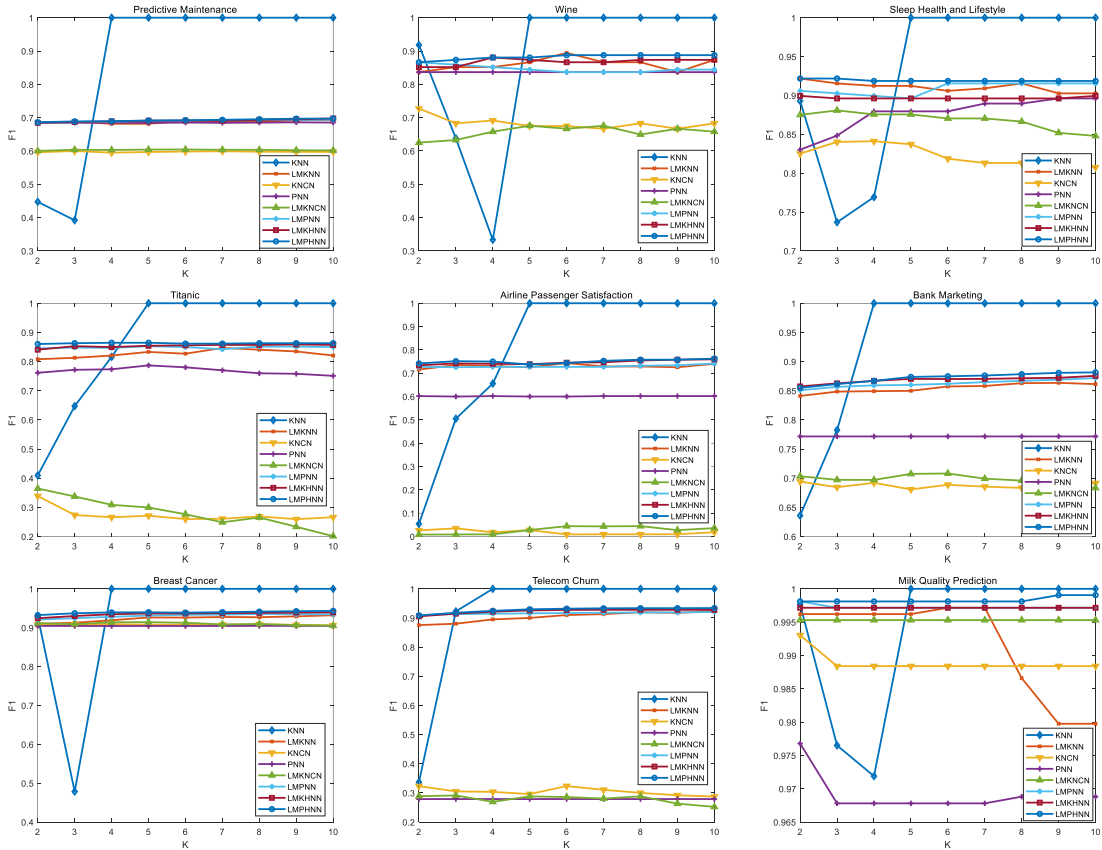

**Figure 4:** F1 Accuracy curve of each model on the datasets

The test set is formed by selecting the remaining samples from each real dataset after randomly choosing the training samples. The ratio of training and test sets is 1:1. We performed a total of nine experiments on each dataset, resulting in nine different training and test sets that were utilized to evaluate performance. In Step 1, we chose the neighborhood parameter k in a range from 2 to 10 for these experiments.

Table 2 displays the experimental findings, it is evident that the proposed LMPHNN classifier consistently outperforms all nine advanced KNN-based methods in terms of classification accuracy across nearly all nine real datasets.

In particular, by introducing the concept of HMD similarity, the LMPHNN classifier focuses on more dependable local means within various classes by coordinating the means, which results in a decreased sensitivity of the proposed method to the parameter k as it more accurately captures the features and similarities between the data and considerably decreases the standard LMPNN rule's error rate, which improves the accuracy and performance of the classification. This part will be proved in 4.3.

For LMKNN, LMKHNN and LMKNCN classifiers, although they have quite high precision or recall on a few datasets, the proposed LMPHNN classifiers are still able to outperform them on most datasets.

In terms of performance, the suggested LMPHNN outperforms the other five methods, according to the results obtained. This is a good indication that our LMPHNN method has better robustness than KNN, LMKNN, LMKNCN, KNCN, LMPNN and PNN.

The following is an analysis of variance results for the data conclusions:

Table 6 Anova results

Anova results (simplified vertical format)

| Class(average±standard deviation) | Accuracy | Recall | Precision | F1 |
| --- | --- | --- | --- | --- |
| KNCN($n$=9) | 0.74±0.14 | 0.56±0.32 | 0.64±0.32 | 0.59±0.32 |
| KNN($n$=9) | 0.74±0.15 | 0.65±0.19 | 0.68±0.17 | 0.66±0.18 |
| LMKHNN($n$=9) | 0.77±0.14 | 0.79±0.13 | 0.97±0.02 | 0.86±0.09 |
| LMKNCN($n$=9) | 0.75±0.15 | 0.56±0.34 | 0.65±0.33 | 0.59±0.33 |
| LMKNN($n$=9) | 0.76±0.14 | 0.78±0.13 | 0.97±0.02 | 0.85±0.10 |
| LMPHNN($n$=9) | 0.78±0.14 | 0.80±0.13 | 0.97±0.02 | 0.87±0.10 |
| LMPNN($n$=9) | 0.76±0.15 | 0.79±0.13 | 0.97±0.02 | 0.86±0.10 |
| PNN($n$=9) | 0.63±0.23 | 0.66±0.21 | 0.95±0.03 | 0.74±0.21 |
| F | 0.854 | 2.154 | 7.664 | 3.522 |
| p | 0.547 | 0.05 | 0.000** | 0.003** |

* $p<0.05$ ** $p<0.01$

The F and p values at the bottom of the table are used to assess the significant differences in the performance of different classifiers.

F-number: A statistic used in ANOVA to compare inter-group and intra-group variances; the larger the F-number, the more significant the difference between groups. P-value: A probability value that tests statistical significance, $p < 0.05$ is generally considered statistically significant, and $p < 0.01$ indicates a very significant difference.

In this table, the p-value of Recall is 0.05, and the P-values of Precision and F1 are 0.000 and 0.003, respectively, indicating that there are significant differences in performance among classifiers on these indicators.

Comparative analysis:

LMPHNN, LMPNN, etc., perform better on Precision and F1 scores because these classifiers have significant P-values (<0.01) on these two indexes.

In contrast, PNN performs poorly in most indexes, especially in Accuracy, with an average value of 0.63.

In terms of performance, based on the results obtained, the proposed LMPHNN is superior to the other five methods. This is a good indication that our LMPHNN method is more robust than KNN, LMKNN, LMKNCN, KNCN, LMPNN, and PNN.

*4.3 Effect of HMD on K-value sensitivity*

To determine the impact of introducing the HMD on the sensitivity of the k value, we conducted an experiment aimed at reducing the sensitivity of the k value to the performance of the algorithm. In this study, we selected three algorithms related to our approach: the LMPNN, the LMKNN, and the LMKHNN, and compared them with our proposed new approach, the LMPHNN. We carried out experiments on a total of four datasets.

To measure the impact of the algorithm's performance on the k values, it is necessary to quantify their sensitivity, and we chose to use the evaluation's standard deviation. Specifically, for each k value, we calculate the differences between all metric values (including accuracy, recall, precision, and F1) and their respective means. Subsequently, we compute the standard deviations of these metric differences. This analysis allows us to assess the stability and sensitivity of the model's performance across different K values. The formula is as follows.

$\sigma$: standard deviation

$k$: the neighborhood size.

$x_i$: represents the value of indicator i. The indicators are accuracy, recall, precision and F1

$\bar{x}$: represents the mean of the indicators, which are accuracy, recall, precision and F1

$$\sigma = \sqrt{\frac{1}{k}\sum_{i=2}^{k}(x_i - \bar{x})^2}$$

(14)

Based on the experiments described above, we conducted a study in the range of K values from 2 to 10. In calculating the standard deviation, we set the value of K to 10, thus obtaining the standard deviation of the indicators with k in the range from 2 to 10.

A smaller standard deviation indicates that the algorithm's results vary to a lesser extent for different values of k, which means that the algorithm is less sensitive to changes in k values. This metric helps us to assess the stability and robustness of the algorithm after the introduction of the HMD, as well as its impact on the sensitivity to k values.

Table 6 Standard deviation of accuracy for each model

| Accuracy | LMPHNN | LMPNN | LMKNN | LMKHNN |
|---|---|---|---|---|
| Wine | **0.0123** | 0.0159 | 0.0285 | 0.0157 |
| SL | **0.0024** | 0.0134 | 0.0108 | 0.0024 |
| Titanic | **0.0021** | 0.0047 | 0.0176 | 0.0079 |
| BC | **0.0054** | 0.0084 | 0.0124 | 0.0081 |

Table 7 Standard deviation of recall for each model

| Recall | LMPHNN | LMPNN | LMKNN | LMKHNN |
|---|---|---|---|---|
| Wine | **0.0339** | 0.0510 | 0.0404 | 0.0426 |
| SL | 0.0270 | **0.0232** | 0.0302 | 0.0347 |
| Titanic | 0.0410 | 0.0412 | **0.0379** | 0.0406 |
| BC | 0.0218 | 0.0212 | 0.0208 | **0.0193** |

Table 8 Standard deviation of precision for each model

| Precision | LMPHNN | LMPNN | LMKNN | LMKHNN |
|---|---|---|---|---|
| Wine | **0.0473** | 0.0534 | 0.0548 | 0.0540 |
| SL | 0.0298 | 0.0355 | **0.0274** | 0.0404 |
| Titanic | **0.0521** | 0.0589 | 0.0666 | 0.0628 |
| BC | **0.0291** | 0.0326 | 0.0369 | 0.0306 |

Table 9 Standard deviation of F1 for each model

| F1 | LMPHNN | LMPNN | LMKNN | LMKHNN |
|---|---|---|---|---|
| Wine | **0.0077** | 0.0105 | 0.0184 | 0.0101 |
| SL | **0.0014** | 0.0080 | 0.0064 | 0.0014 |
| Titanic | **0.0013** | 0.0031 | 0.0121 | 0.0053 |
| BC | **0.0030** | 0.0048 | 0.0072 | 0.0046 |

Based on the results in the above table, it can be clearly seen that LMPHNN performs well in terms of overall performance. This not only provides solid support for our new approach but also further validates the positive impact of the introduction of HMD in producing a significant reduction in the sensitivity of the classifier to k-value changes. This finding provides strong evidence that our algorithm is able to achieve better performance under different k-values while maintaining stability and robustness, which is of great practical significance for solving real-world problems in related fields.

**5 Limitation analysis due to assumptions and other factors**

As a classifier algorithm, LMPHNN has the following limitations caused by assumptions:

1. The "curse of dimension" faced by the traditional KNN algorithm will also make the distance of the algorithm less meaningful in the case of high dimensions, but the LMPHNN algorithm adopts HMD and local mean to alleviate this problem to the greatest extent

2. In high-dimensional space, the accuracy and representativeness of LMV decrease slightly due to the sparse distribution of nearest neighbors

3. In the hypothesis, we believe that the local mean vector can effectively represent a category, but, if the important features within the category are extremely rich, LMV may not be able to fully capture such diversity

4. The computational complexity increases relative to the basic KNN algorithm

5. Although the influence of k value and abnormal noise is reduced to the maximum extent, the performance of the classifier will be affected if the situation is more extreme

In addition to the above effects that ordinary data sets may have, for small sample data sets, although we have proved in principle and experiment that it can optimize the training results of small samples to the greatest extent, it will still be affected in the following aspects:

1. Because there are few sample categories and a small amount of data, it is easy to overfit the training data, resulting in model errors

2. In the case of small samples, few data of some categories may cause the local mean vector to be difficult to represent the characteristics of the category data

3. If the data features are not independently and equally distributed, and have quite complex distribution and structure, there will be great obstacles in the extraction of data features from small samples

**6 Performance comparison with sota model**

For the sota model, this paper selects "kNN-P: A kNN classifier optimized by P systems", published in the journal Theoretical Computer Science in 2020, proposes a P-system-based K-nearest neighbor (kNN) classifier optimization method. It's called kNN-P. This approach takes advantage of the concept of membrane computing, a distributed, parallel computing model inspired by the structure and function of biological cells. In kNN-P, a P system consists of multiple cells, each of which is responsible for determining an optimal set of k nearest neighbors for a test sample. By controlling the rules of evolution and communication, each cell can independently search and optimize its solution set.

Specifically, the main features of the kNN-P algorithm include:

1. Membrane computing framework: The P system provides a computing framework for simulating the parallel processing capabilities of cells.

2. Optimize k value: The algorithm tries to find the optimal k value, that is, the number of nearest neighbors, which is a key parameter in the kNN algorithm.

3. Parallel processing: By processing test samples in parallel in multiple cells of the P system, the efficiency of the algorithm is improved.

4. Communication mechanisms: Information is exchanged between cells through communication mechanisms to update and improve their respective solution sets.

5. Fitness function: The solution in each cell is evaluated by the fitness function to determine its effectiveness in classifying the test sample.

6. Evolutionary rules: Similar to the rules in particle swarm optimization (PSO), the solutions in the cell are updated according to fitness.

The following is the pseudo-code section:

```
BEGIN
    CLEAR all previous data
    CLEAR console
    LOAD dataset from specified file
    EXTRACT labels (last column) from the dataset
    REMOVE the last column from the dataset (features only)
    SET train_ratio to 0.7
    SET test_ratio to 0.3
    REMOVE classes that are too small to appear in the test set
    PARTITION data into training and testing sets using cvpartition
    GET training indices and testing indices
    EXTRACT training data and training labels using training indices
    EXTRACT testing data and testing labels using testing indices
    INITIALIZE parameters:
        q = number of test samples
        k = number of nearest neighbors (e.g., 5)
        maxstep = maximum number of iterations (e.g., 100)
        w, c1, c2 = PSO parameters (though not used in simplified version)
    INITIALIZE particles (k nearest neighbors for each test sample)
    INITIALIZE velocities (not used in simplified version)
    FOR each iteration from 1 to maxstep:
        FOR each test sample i:
            COMPUTE distances between the i-th test sample and all training samples
            SORT distances to find the indices of k nearest neighbors
            STORE indices of k nearest neighbors in particles
    INITIALIZE predictions array
    FOR each test sample i:
        GET labels of k nearest neighbors from training labels
        DETERMINE the majority class label (mode) from neighbors' labels
        STORE the predicted class for the i-th test sample in predictions
        PRINT classification result for the i-th test sample
    COMPUTE confusion matrix using true labels and predicted labels
    CALCULATE number of true positives, total samples, false positives, false negatives, and true negatives
    CALCULATE accuracy, error rate, precision, recall, and F1 score
    HANDLE cases where precision or recall are NaN
    STORE results in a result table
    DISPLAY the result table
END
```

After testing, under the same data set, the accuracy rate is 0.5510, the recall rate is 0.7555, the accuracy rate is 0.6706, and the F1 score is 0.7105. While the average value of various indicators of LMPHNN measured repeatedly under the same data set is as follows: With an error rate of 0.1567, an accuracy rate of 0.8433, a recall rate of 0.9179, an accuracy rate of 0.9122, and an F1 score of 0.9150, LMPHNN is the overall leader in average data performance for this open dataset.

**7 Conclusions**

This paper presents a newly developed technique referred to as the pseudo-nearest neighbor rule based on the local mean of the harmonic mean distance(LMPHNN). Our inspiration for LMPHNN comes from the LMPNN and HMD rules. LMPHNN serves as an extension of LMPNN with the primary objective of addressing outliers to enhance classification performance. Within the framework of LMPHNN, we employ the HMD as a similarity metric and leverage LMVs from the LMPNN rule for classification based on the respective LMVs of each nearest neighbor within each class. Subsequently, the assigned class label to the query pattern is determined based on the class label of the PNN that is nearest, as determined by the local mean. To judge the classification performance of this approach, extensive experiments were conducted across nine real UCI and Kaggle datasets. Comparative analyses were performed against well-known classifiers including KNN, LMKNN, LMKNCN, KNCN, LMPHNN, and PNN. The consistent experimental outcomes underscore the effectiveness and stability of the LMPHNN method, resulting in favorable classification performance. Furthermore, an area for future research lies in the development of an adaptive LMPNN method that incorporates the harmonic mean distance for pattern classification.

**8 The development and discussion of the KNN classifier**

The core of the KNN classifier lies in the selection of distance metric, and the research progress in recent years mainly focuses on the following directions:

1. Optimization of distance measurement: Researchers have explored a variety of distance measurement methods, such as Minkowski distance and Mahalanobis distance, to adapt to the characteristics of different data sets.

2. Entropy feature transformation: Entropy feature transformation reduces the class noise of feature parameters and improves the classification accuracy of the KNN algorithm.

3. Deep learning fusion: Combined with convolutional neural network (CNN) to learn discriminant features and distance metrics to improve the performance of the KNN algorithm.

4. Multi-scale distance measurement: capture the similarity of data at different scales to enhance the generalization ability of the algorithm.

5. Ensemble learning application: Build the integration of multiple KNN classifiers and improve the classification performance through the integration method.

6. Combination of clustering algorithms: such as density peak clustering algorithm, auxiliary KNN algorithm classification decision.

7. Adaptive distance adjustment: Develop distance measurement methods for adaptive data characteristics and dynamically adjust parameters to adapt to different data sets.

8. Large-scale data processing: Optimization strategies are proposed to improve the efficiency and accuracy of the KNN algorithm on large-scale data sets.

These advances demonstrate the potential of KNN algorithms for handling complex data sets and improving classification accuracy. Future research may focus more on the scalability of LMPHNN algorithms and their integration with other technologies.

**9 Potential fields and directions for the future**

Practical application discussion:

CMC, 202x, vol.xx, no.xx                                                                                                xxxx1. Medical diagnosis: LMPHNN can be applied to medical data analysis to help doctors quickly diagnose diseases based on a small number of patient signs and symptoms, especially in the case of rare diseases or emerging diseases.

2. Financial fraud detection: In the financial field, the number of abnormal transactions tends to be small, LMP.

3. Network security: LMPHNN can be used to detect network intrusion and abnormal traffic in real-time, especially in the early stage of the attack when the abnormal behavior samples are small.

Expansion of future research directions:

1. Adapt to unbalanced data sets: Study how to adjust the LMPHNN algorithm to better handle unbalanced data sets, such as through sampling techniques or reweighting methods.

2. Ensemble learning methods: Explore ensemble learning frameworks that combine LMPHNN with other machine learning algorithms, such as Boosting or Bagging, to improve overall classification performance.

3. Deep learning integration: Consider combining LMPHNN with a deep learning model, making use of the advantages of deep learning in feature extraction, and using LMPHNN to make the final classification decision.

4. Multimodal learning: Study the application of LMPHNN on multimodal data, such as combining text, image, and sound data for classification.

5. Real-time classification systems: LMPHNN algorithms are optimized to meet the needs of real-time data processing, such as applications on Internet of Things (IoT) devices.

6. Cross-field adaptability: Study the adaptability and effectiveness of LMPHNN in different fields (such as bioinformatics, environmental science, etc.)

**Funding Statement:** The authors received no specific funding for this study.

**Availability of Data and Materials:** The data that support the findings of this study are available from the corresponding author, Z. Lu, upon reasonable request.

**Conflicts of Interest:** The authors declare that they have no conflicts of interest to report regarding the present study.**References**

[1]     T. Cover and P. Hart, "Nearest neighbor pattern classification," *IEEE Trans. Inf. Theory*, vol. 13, no. 1, pp. 21–27, Jan. 1967.

[2]     Z. Ma, Z. Liu, C. Luo, and L. Song, "Evidential classification of incomplete instance based on K-nearest centroid neighbor," *J. Intell. Fuzzy Syst.*, vol. 41, no. 6, pp. 7101–7115, Jan. 2021.

[3]     Y. Mitani and Y. Hamamoto, "A local mean-based nonparametric classifier," *Pattern Recognit. Lett.*, vol. 27, no. 10, pp. 1151–1159, Jul. 2006.

[4]     J. Gou, Z. Yi, L. Du, and T. Xiong, "A Local Mean-Based k-Nearest Centroid Neighbor Classifier," *Comput. J.*, vol. 55, no. 9, pp. 1058–1071, Sep. 2012.

[5]     E. Fix and J. L. Hodges, "Discriminatory Analysis. Nonparametric Discrimination: Consistency Properties," *Int. Stat. Rev. Rev. Int. Stat.*, vol. 57, no. 3, pp. 238–247, 1989.

[6]     T. Wagner, "Convergence of the nearest neighbor rule," *IEEE Trans. Inf. Theory*, vol. 17, no. 5, pp. 566–571, Sep. 1971.

[7]     X. Wu *et al.*, "Top 10 algorithms in data mining," *Knowl. Inf. Syst.*, vol. 14, no. 1, pp. 1–37, Jan. 2008.

[8]     J. Gou, L. Du, Y. Zhang, and T. Xiong, "A New Distance-weighted k -nearest Neighbor Classifier," *J Inf Comput Sci*, vol. 9, Nov. 2011.